\documentclass{article}

\usepackage{microtype}
\usepackage{graphicx}
\usepackage{subcaption}
\usepackage{booktabs} 
\usepackage{amsmath}

\usepackage{hyperref}

\usepackage[preprint]{icml2026}

\usepackage{amssymb}
\usepackage{mathtools}
\usepackage{amsthm}
\usepackage{tabularx}
\usepackage{longtable,booktabs}
\usepackage{multirow}

\usepackage{algorithm}
\usepackage{algorithmic}

\usepackage[capitalize,noabbrev]{cleveref}

\theoremstyle{plain}

\theoremstyle{definition}

\theoremstyle{remark}

\usepackage[textsize=tiny]{todonotes}

\icmltitlerunning{ExecVLA: Following Fine-Grained Execution Constraints in Vision-Language-Action Models with Bi-Level Action Representation}

\begin{document}

\twocolumn[
  \icmltitle{ExecVLA: Following Fine-Grained Execution Constraints in Vision-Language-Action Models with Bi-Level Action Representation}



  \icmlsetsymbol{equal}{*}

  \begin{icmlauthorlist}
 \icmlauthor{Gaoge Han, Zhengqing Gao, Ziwen Li, Jiaxin Huang, Shaoli Huang, Fakhri Karray, Mingming Gong, Tongliang Liu}{a}
  
  \end{icmlauthorlist}

  \icmlaffiliation{a}{MBZUAI}

  \icmlcorrespondingauthor{Tongliang Liu}{tliang.liu@gmail.com}

  \icmlkeywords{VLA}

  \vskip 0.3in
]



\printAffiliationsAndNotice{}  

\begin{abstract}

We study fine-grained execution-constraint following in vision-language-action (VLA) models. Given an invariant task goal, the policy must follow instruction-specified execution constraints, including interaction targets, motion patterns, spatial relations, and terminal configurations. This setting exposes a limitation of goal-oriented VLAs: trajectories that complete the same task are not interchangeable when the instruction specifies how the task must be executed. We propose ExecVLA, a framework that separates a goal-oriented component from an execution-specific component through a bi-level action representation and supervised bi-level reasoning tokens. We further introduce explicit goal-invariance and execution-predictability objectives so that the goal-level representation remains stable across executions of the same goal, while the execution-level representation retains the constraints that distinguish those executions. We construct execution-constraint-following datasets in simulation and on a Realman-75 robot, with goal and fine-grained reasoning annotations. Experiments on LIBERO and the real robot show improved goal completion and, more importantly, substantially more reliable adherence to instruction-specified execution constraints.

\end{abstract}

\section{Introduction}
\label{sec:intro}

\begin{figure}
    \centering
    \includegraphics[width=0.95\linewidth]{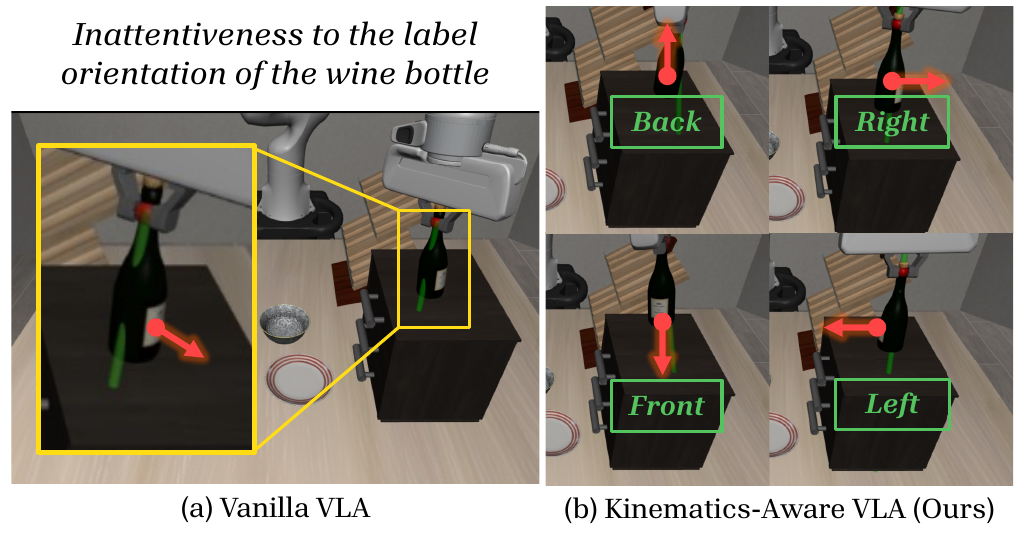}
    \caption{Comparison of Vanilla VLAs vs. KineVLA. Vanilla VLAs~\cite{kim2025openvla,wang2025vq}, which accept coarse command inputs (e.g., ``place the wine bottle on the drawer") and produce relatively fixed bottle label orientation actions. In contrast, Our KineVLA can processes fine kinematic commands (e.g., ``control the wine bottle to face a specific orientation on the cabinet") and generate robot end-effector actions oriented \emph{front, back, left, or right.}}
    \label{fig:teaser}
\end{figure}

Vision-Language-Action (VLA) models have recently demonstrated strong performance in goal-oriented robotic manipulation. However, most existing approaches implicitly assume goal equivalence under instruction paraphrasing, meaning that instructions describing the same task goal can be executed with interchangeable motion trajectories. As a result, these models are largely insensitive to fine-grained kinematic specifications expressed in natural language. In real-world human robot interaction, this assumption frequently breaks down. Semantically equivalent goals may require distinct and non-interchangeable trajectories due to explicit kinematic constraints such as position, orientation, distance, and motion direction.

For example, as shown in Figure~\ref{fig:teaser}, instructions such as “place the bottle on the drawer” and “place the bottle on the drawer with the label facing right, front, or back” describe an identical task goal, yet they demand fundamentally different execution trajectories and end-effector motions. These trajectories cannot be treated as interchangeable refinements of the same action without violating the instruction semantics. This setting goes beyond conventional fine-grained goal specification and instead characterizes a class of tasks that are goal invariant but kinematics variant. Such tasks are not captured by the implicit assumptions underlying most existing VLA formulations.

\begin{figure*}
    \centering
    \includegraphics[width=0.81\linewidth]{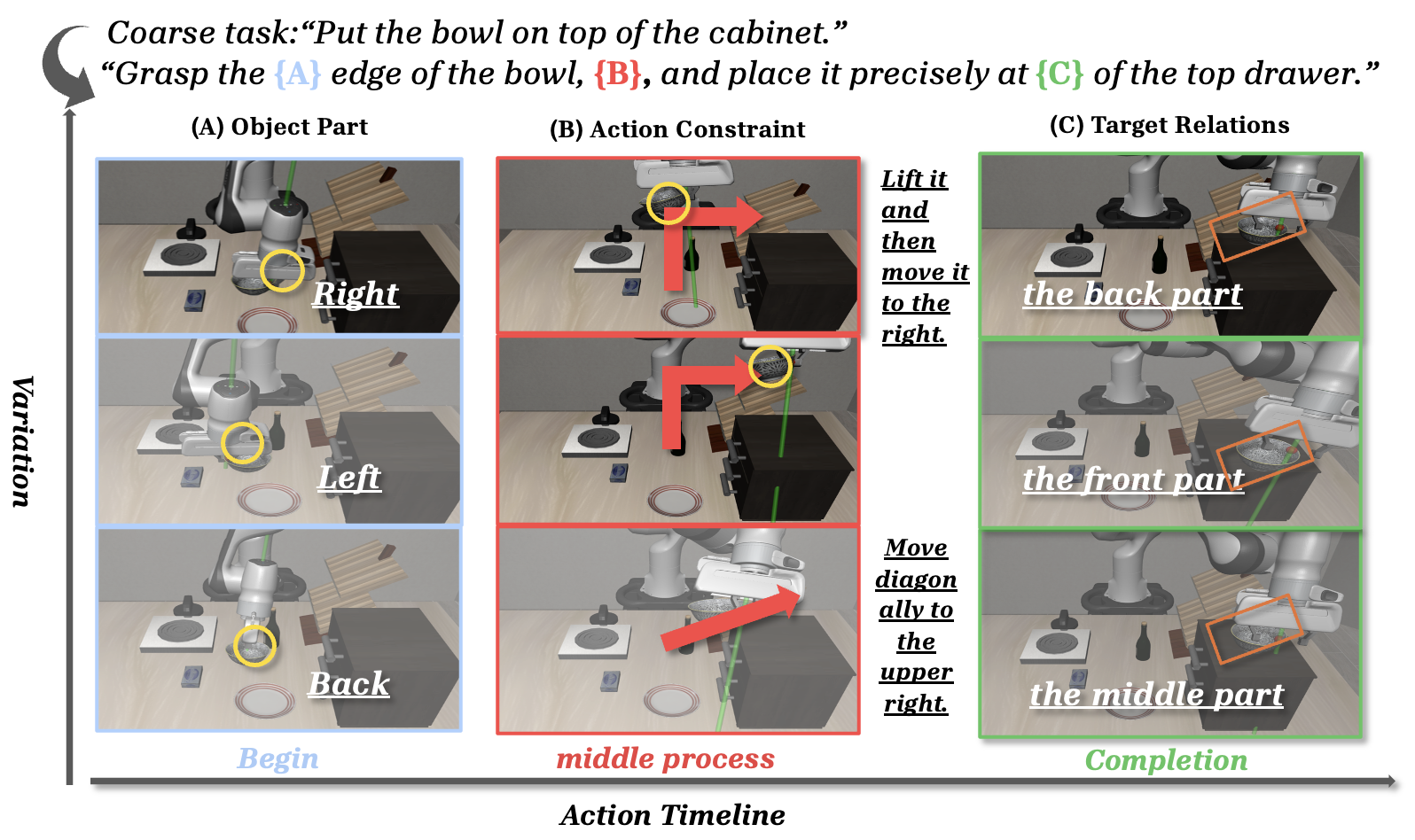}
    \caption{A task example from our proposed \textbf{Kinematics-Rich Datasets}: In contrast to coarse task instructions that aim to complete a general goal, our approach captures diverse, fine-grained kinematic variations for an action instruction and their temporal evolution across multiple key action stages. These variations encompass \{A\} Object Part, \{B\} Action Constraint, and \{C\} Target Relations, with corresponding images of key actions displayed from left to right. Our KineVLA method is designed to address this challenge, excelling at perceiving these multi-faceted details to achieve precise manipulation.}
    \label{fig:teaser_data}
\end{figure*}

Despite the rapid progress in recent VLA models such as OpenVLA~\cite{kim2025openvla} and $\pi_0$~\cite{black2024pi_0}, most existing approaches focus on semantic-level grounding, mapping visual observations and language descriptions to discrete action tokens or low-frequency control trajectories. While these models generalize well to diverse manipulation tasks, their generated actions typically reflect task goal rather than motion precision. In other words, current VLAs are optimized to determine what to do (e.g., grasp or place) but not how to do it (e.g., at which position, distance, or orientation). Moreover, although tokenization-based or diffusion-based decoders improve temporal smoothness, they remain agnostic to fine-grained kinematic cues in language. This gap prevents existing policies from achieving physically consistent, user-customized manipulations that depend on explicit kinematic instructions.

Naive binning struggles with complex actions, while diffusion-based policies are computationally expensive. We therefore adopt vector-quantized action representations~\cite{lee2024behavior,wang2025vq,mete2024quest,pertsch2025fast} to capture fine-grained motion and temporal consistency for more stable, sample efficient training. To avoid the entanglement of task identity and kinematic realization in single-level tokenizers, we further introduce a bi-level action representation that decouples goals and kinematics, making it well suited for kinematics-rich tasks.

In order to achieve greater sensitivity to sophisticated fine-grained kinematic actions while preserving robustness to coarse-grained actions, KineVLA proposes a bi-level vector-quantized action discretization scheme, which decomposes robot actions into two complementary latent spaces: a goal-level codebook that captures semantic goals and task goal, and a kinematics-level codebook that encodes precise motion parameters such as direction, distance, and velocity. Bi-level action tokens structure the action space but do not by themselves guarantee correct grounding of kinematic language. We therefore introduce bi-level reasoning tokens as explicit intermediate variables to align instruction-level constraints with hierarchical action representations. Bi-level reasoning tokens parse the task goal and kinematic details of the command. Thus resulting in a bi-level chain-of-thought (CoT)-style generation paradigm, in which the model jointly generates textual reasoning (\textit{Reasoning}) and discrete action tokens (\textit{Action}). To ensure consistency between reasoning and control, we maximize the conditional mutual information $I(\textit{Reasoning}; \textit{Action}\mid C)$, which encourages the generated trajectory to remain semantically grounded, thus forming a coherent link between reasoning and execution. Our reasoning tokens are not intended to improve abstract planning ability, but to serve as explicit, supervised intermediate variables that align language and action.

To alleviate the data scarcity of this new task, we constructed three \textbf{Kinematics-Rich Datasets} that systematically captures a variety of goals and multi-level kinematic details, covering diverse tabletop organization and object manipulation scenarios. The total number of frames are 378K. We collected paired VLA data using the LIBERO platform~\cite{liu2023libero} and the Realman-75 robot. The kinematics-rich language instructions are also annotated with reasoning texts at two levels: 1) general coarse-grained goal, and 2) fine-grained kinematics. Our kinematics-rich VLA datasets capture the entire process of action execution, not only coarsely or partially, as illustrated in Figure~\ref{fig:teaser_data}. The dataset encompasses diverse actions that encode object-part interactions, target spatial relations, and kinematic action constraints. Such a detailed annotation scheme enables the model to learn highly precise motion representations.

Extensive experiments show that KineVLA achieves state-of-the-art performance on kinematics-aware benchmarks. Unlike previous models that produce relatively rigid, pre-defined, and goal-directed motions, our model performs flexible, interpretable, and kinematics-aware operations directly from kinematics-rich instructions.

In summary, our main contributions are as follows:
\begin{itemize}
    \item We introduce a new  kinematic-rich manipulation task, which bridges semantic goals with diverse action execution trajectories, with the aim of meeting the needs of users’ for customized table organization tasks.
    \item We present \textbf{KineVLA}, a VLA model with bi-level vector quantized action representation that decouples task goals from kinematic variations, enabling interpretable and kinematically sensitive control.
    \item We construct three new \textbf{Kinematics-Aware Manipulation Dataset} covering both simulated and real-world scenarios, with detailed spatial, object-level, and motion-constraint annotations to facilitate kinematic reasoning.
\end{itemize}

\section{Related Work}
\label{sec:related}
\subsection{Vision-Language-Action Models}
In recent years, VLA models have rapidly progressed from early end-to-end robotic policies grounded in paired visual perception and text to generalist systems capable of long-horizon, multi-stage manipulation. Works such as RT-2~\cite{zitkovich2023rt} and RT-X~\cite{o2024open} demonstrated that grounding large vision–language models in real robot trajectories can yield surprisingly strong zero-shot control across tasks and embodiments. OpenVLA~\cite{kim2025openvla} established an influential open-source baseline by coupling a 7B LLM with strong visual encoders and nearly a million real robot demonstrations, showing competitive, cross-embodiment generalization, and efficient task adaptation. More recently, the $\pi_0$~\cite{black2024pi_0} family reframed action generation with flow-matching and high-frequency action chunking for dexterous control, yielding robust closed-loop visuomotor skills from mixed supervisory signals. Building on this, $\pi_{0.5}$~\cite{intelligence2025pi_05} co-trains across heterogeneous data sources—multi-robot logs, web-scale semantics, and auxiliary prediction tasks to widen open-world generalization in the wild.

Based on their action decoding strategies, current VLA models can be broadly categorized into three classes: those employing per-timestep binning scheme (e.g., ~\citep{brohan2022rt,zitkovich2023rt,kim2025openvla}), diffusion-based decoding (e.g., ~\citep{liu2024rdt,liu2025hybridvla,black2024pi_0,intelligence2025pi_05}), and those adopting vector-quantized action
representations (e.g., ~\citep{belkhaleminivla,lee2024behavior,mete2024quest,pertsch2025fast,wang2025vq}). In our work, we emphasize the structural nature of action primitives and their compositional semantics. To this end, we adopt vector-quantized action representations approach and introduce a bi-level RVQ-VAE training framework, which jointly captures global task context and fine-grained kinematic patterns, enabling temporally coherent and semantically grounded action decoding.

\subsection{Physically Reliable Robot Manipulation}
As the community increasingly seeks to enhance robots’ ability to interact reliably with the real physical world, physically plausible robot manipulation has emerged as a key research frontier. Instead of relying solely on visual imitation or simulation priors, recent works integrate force–torque sensing~\cite{yu2025forcevla,he2025foar,huang2025tactile}, tactile feedback~\cite{huang2025tactile,wu2025freetacman,bi2025vla}, and affordance-guided~\cite{xu2024naturalvlm} reasoning to better capture the dynamics of contact and object interaction.

ForceVLA~\citep{yu2025forcevla} and FoAR~\citep{he2025foar} augment VLA pipelines with wrench data and impedance feedback, improving compliance, force regulation, and fine contact control in real manipulation. Tactile-VLA~\citep{huang2025tactile} and VLA-Touch~\citep{bi2025vla} demonstrate the benefit of integrating high-frequency tactile embeddings and visuotactile fusion, enabling policies that can infer slip, deformation, and surface geometry directly from touch signals. FreeTacMan~\citep{wu2025freetacman} further scales tactile-conditioned manipulation across diverse objects and sensors through self-supervised visuotactile representation learning. In parallel, affordance-driven models such as NaturalVLM~\citep{xu2024naturalvlm} bridge semantic perception and physical reasoning by predicting object parts, functional regions, and graspable affordances conditioned on natural-language queries.

Overall, these efforts can be viewed as emphasizing physical dynamics through the incorporation of additional sensory modalities. In contrast, our work focuses on fine-grained, user-customized instruction tasks, highlighting kinematic-level reasoning of robot motion under diverse physics-aware settings, thereby bridging linguistic goal and physically grounded, precise action execution.

\section{Method}
\label{sec:method}
In this section, we present our method, comprising a bi-level RVQ-VAE and KineVLA. We begin by presenting bi-level vector quantized action representation (Sec.~\ref{subsec:bilevel-vqvae}), detailing its implementation specifics of training a Bi-Level Residual VQ-VQE. We then describe the bi-level generation paradigm and the associated mutual-information regularization scheme (Sec.~\ref{subsec:cot-generation}).

\begin{figure*}
    \centering
    \includegraphics[width=0.9\linewidth]{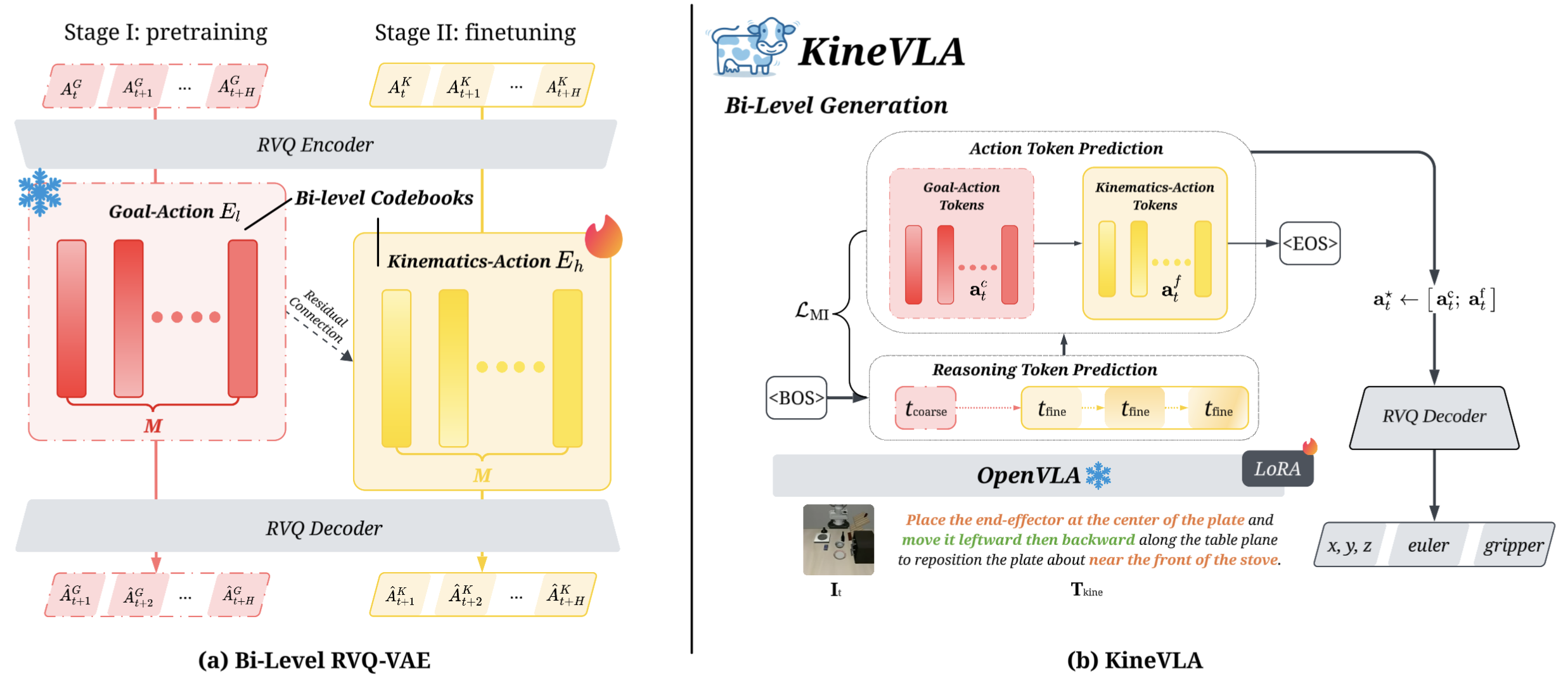} 
    \caption{Overview of our framework. Our approach decouples low-frequency, goal-level control from fine-grained kinematic refinements to effectively handle kinematics-rich tasks. (a) The proposed Bi-Level RVQ-VAE learns hierarchical action representations (Sec.~\ref{subsec:bilevel-vqvae}), while (b) the KineVLA framework addresses kinematics-rich tasks through bi-level generation (Sec.~\ref{subsec:cot-generation}).}
    \label{fig:pipline}
\end{figure*}

\subsection{Bi-Level Action Representation}
\label{subsec:bilevel-vqvae}

Uniform binning of each action dimension, as in OpenVLA-style discretization where token IDs lie in $[0,255]$, fixes partition boundaries and ignores cross-joint correlations. This affects the robustness of low-frequency trends and sensitivity to high-frequency transients. We instead learn discrete \emph{vector-quantized} (VQ) action tokens inspired by~\cite{zeghidour2021soundstream,lee2024behavior,wang2025vq}, then extend them to a \emph{bi-level} design that targets both regimes.

\noindent \textbf{Vector-Quantized Action Representation.}
Given a chunk of an action sequence $A_{t:t+H}\!\in\!\mathbb{R}^{T\times D}$, an encoder $f_\theta$ produces latents $z_t=f_\theta(A_{t:t+H})\in\mathbb{R}^d$. It uses $M$ codebooks $E^{(m)}=\{e^{(m)}_k\}_{k=1}^{K_m}$ applied sequentially to quantify residuals by Residual Vector Quantized Variational AutoEncoder (RVQ-VAE)~\cite{zeghidour2021soundstream}. Initialize $r_t^{(0)}=z_t$ and for $m=1,\dots,M$:
\begin{equation}
\begin{aligned}
k_t^{(m)} &= \arg\min_k \big\lVert r_t^{(m-1)} - e_k^{(m)} \big\rVert_2^2, \\
\tilde z_t^{(m)} &= e_{k_t^{(m)}}^{(m)}, \\
r_t^{(m)} &= r_t^{(m-1)} - \tilde z_t^{(m)}.
\end{aligned}
\end{equation}
The final quantized latent and discrete token tuple are
\begin{equation}
\tilde z_t=\sum_{m=1}^M \tilde z_t^{(m)},\qquad
\mathbf{i}_t=\big(k_t^{(1)},\dots,k_t^{(M)}\big) \text{ (tokens)}.
\end{equation}
A decoder $g_\phi$ reconstructs continuous actions $\hat A_{t:t+H}=g_\phi(\tilde z_{1:T})$. We train with a reconstruction term and RVQ codebook and commitment terms:
\begin{equation}
\begin{aligned}
\mathcal{L}_{\text{rec}} &= \frac{1}{T}\sum_{t=1}^T \lVert A_t-\hat A_t\rVert_1,\\[0.25em]
\mathcal{L}_{\text{rvq}} &= \mathcal{L}_{\text{rec}}
+ \sum_{m=1}^M \sum_{t=1}^T 
   \big\lVert \mathrm{sg}(r_t^{(m-1)}) - \tilde z_t^{(m)} \big\rVert_2^2 \\
&\quad + \sum_{m=1}^M \sum_{t=1}^T 
   \beta_m \big\lVert r_t^{(m-1)} - \mathrm{sg}(\tilde z_t^{(m)}) \big\rVert_2^2 ,
\end{aligned}
\end{equation}
where $\mathrm{sg}(\cdot)$ is the stop-gradient operator and $\beta_m$ balances commitment at level $m$. Unlike OpenVLA's binning scheme, our partitions are learned, non-uniform, and compositional across $M$ residual levels, which couples joints and time while yielding compact semantic tokens.

\noindent \textbf{Bi-Level Action Representation.}
Building on the above vector-quantized action representation formulation, we factor actions into two discrete spaces backed by two codebooks of identical capacity: a goal (coarse-grained) level that captures low-frequency structure and task goal, and a kinematics (fine-grained) level that captures high-frequency corrections and contact-rich transients. Let $E_l\in\mathbb{R}^{K\times d}$ and $E_h\in\mathbb{R}^{K\times d}$ denote the goal and kinematics codebooks (same as $K,d$), which are then combined and fed to the decoder to reconstruct the action chunk.

\noindent \textbf{Bi-Level RVQ-VAE Training.}
We adopt a two-stage training schedule with stepwise codebook learning, while keeping the two codebooks of equal size. This design explicitly encourages the fine-grained codebook to capture kinematics-sensitive action representations, enabling it to model subtle action variations more effectively:

\noindent \textbf{Stage I (pretraining with coarse-grained actions $A^G_{t:t+H}$).} Train the coarse path with RVQ-VAE on broad, multi-task data (e.g., Open X-Embodiment~\cite{o2024open}, ManiSkill~\cite{mu2021maniskill}) to learn $E_l$ and its quantizer. This stage encourages temporally stable, low-frequency tokens.

\noindent \textbf{Stage II (finetuning with kinematics-aware actions $A^K_{t:t+H}$).} Freeze $E_l$ and the coarse-grained codebook quantizer. Initialize a residual path with a new RVQ stack and codebook $E_h$ of the same size, warm-start the shared encoder/decoder from Stage I, and train on \emph{Kine Datasets} to specialize $E_h$ for high-frequency details. Only $E_h$ and the fine-grained codebook quantizer are learned from scratch; shared modules are finetuned.

\subsection{Bi-Level Generation}
\label{subsec:cot-generation}

After decoupling actions into a two-level representation, we predict bi-level reasoning tokens to align internal representations with language parsing. This is inspired by chain-of-thought in large language models~\cite{yang2025qwen3,guo2025deepseek}, this design improves interpretability, constraint injection and is trained end to end via supervised finetuning, avoiding the instability of reinforcement learning~\cite{yu2025dapo}. To improve coherence between reasoning and action tokens, we introduce a mutual information regularization that aligns the two bi-level streams and reduces execution drift.

\noindent\textbf{Reasoning Tokens Prediction.}

Given the input image $\mathbf{I_t}$ and a kinematics rich instruction $\mathbf{T}_{\text{kine}}$, we fine-tuning an end to end base VLM model $f_{\theta}$ to produce two reasoning text outputs $t_\mathrm{coarse}$ and $t_\mathrm{fine}$. The coarse reasoning text expresses the general task goal in natural language, while the fine reasoning text specifies key kinematics parameters and anchor points; see Section.~\ref{sec:experi} for concrete examples. Training uses a language cross entropy objective over both outputs,
\begin{equation}
\small
\mathcal{L}_{reasoning}(\theta)
=\sum_{m\in\{c,f\}}\sum_{t=1}^{L_{m}}
\mathrm{CE}\!\Big(h^{m}_{t},\, p_{\theta}\!\big(h^{m}_{t}\mid h^{m}_{<t},\, \mathcal{C}\big)\Big),
\end{equation}

\noindent where $\mathcal{C}=(\mathbf{I_t},\mathbf{t}_{\text{kine}})$, \(h^{c}_{t}\) and \(h^{f}_{t}\) are tokens from the coarse and fine reasoning texts, \(L_{c}\) and \(L_{f}\) are their lengths.

\noindent \textbf{Action Tokens Prediction.}
We use the bi-level action representation derived in Sec.~\ref{subsec:bilevel-vqvae} as ground-truth labels for supervised prediction. Because the action codebook indices learned by the bi-level RVQ-VAE reside in their own index space, we offset these indices to the tail of the text vocabulary. This lets the model reuse a shared output head and apply a unified loss across both text and action tokens, avoiding any extra task-specific head while keeping supervision consistent. Following the reasoning text prediction formulation, the prediction objective is
\begin{equation}
\small
\mathcal{L}_{action}(\theta)
=\sum_{m\in\{c,f\}}\sum_{t=1}^{L_{m}}
\mathrm{CE}\!\Big(a^{m}_{t},\, p_{\theta}\!\big(a^{m}_{t}\mid a^{m}_{<t},\, \mathcal{C}\big)\Big),
\end{equation}

\noindent where \(a^{c}_{t}\) and \(a^{f}_{t}\) are tokens from the coarse-grained and fine-grained action tokens.

\noindent \textbf{Mutual Information Regularization for reason Text and Action.}
Although the proposed model jointly generates textual reasoning (\textit{Reasoning}) and continuous control trajectories (\textit{Action}), 
these two objectives are often optimized independently, which can lead to semantic disconnection. 
In such cases, textual reasoning may not faithfully reflect the underlying motor intention. 
To address this issue, we maximize conditional mutual information 
$I(\text{Reasoning}; \text{Action} \mid C)$, where $C$ denotes the given context (e.g., task condition or observation). 
We achieve this by pulling paired embeddings $(\text{Reasoning}, \text{Action})$ closer together 
and pushing mismatched pairs apart using an InfoNCE objective~\cite{oord2018representation,he2020momentum,chen2020simple}.

Formally, given a batch $\{(T_i, A_i, C_i)\}_{i=1}^N$, 
we obtain embeddings 
$\mathbf{t}_i = f_T(T_i, C_i)$ and $\mathbf{a}_i = f_A(A_i, C_i)$, 
and define the similarity
\begin{equation}
s_{ij} = \frac{\mathbf{t}_i^\top \mathbf{a}_j}{\tau},
\label{eq:sim}
\end{equation}
where $\tau > 0$ is a temperature parameter.

The symmetric mutual information regularization loss is
\begin{equation}
\small
\mathcal{L}_{\text{MI}}
=
-\frac{1}{2N}\sum_{i=1}^N
\left[
\log\frac{e^{s_{ii}}}{\sum_{j=1}^N e^{s_{ij}}}
+
\log\frac{e^{s_{ii}}}{\sum_{j=1}^N e^{s_{ji}}}
\right].
\label{eq:mi_symmetric}
\end{equation}

\noindent \textbf{Bi-Level Action Coupling Inference.}
As presented in Algorithm~\ref{alg:bi_level_cot} , we perform action inference from a scene image, a wrist image, and a kinematic instruction using a generator--decoder pipeline. 
The generator emits bi-level chain-of-thought text and discrete action codebooks: a high-level plan and a low-level control trace. 
Each level is delimited with special tokens and paired with its own action codes. 
A bi-level decoder then fuses both codebooks to reconstruct a single action chunk.

\begin{algorithm}[t]
\caption{Bi-Level Action Coupling Inference}
\label{alg:bi_level_cot}
\begin{algorithmic}[1]
\REQUIRE Sequence length $T$, chunk size $H{=}5$;
         observations $\{(\mathbf{I}_t,\mathbf{W}_t)\}_{t=1}^T$;
         kinematic instruction $\mathbf{T}_{\text{kine}}$;
         generator $\mathcal{G}$;
         bi-level decoder $\mathcal{D}$ 
\ENSURE Predicted action chunks $\{\hat{\mathbf{A}}_{t:t+H}\}_{t=1}^{T-H}$

\FOR{$t = 1$ \textbf{to} $T-H$}
    \STATE $\mathbf{o}_t \gets \mathrm{Enc}_{\text{obs}}(\mathbf{I}_t, \mathbf{W}_t)$
    \STATE $\mathbf{t}_{\text{kine}} \gets \mathrm{Enc}_{\text{kin}}(\mathbf{T}_{\text{kine}})$
    \STATE $\mathcal{C}_t \gets \mathrm{BuildContext}(\mathbf{o}_t, \mathbf{t}_{\text{kine}})$ 
    \STATE $(\mathbf{h}^{\mathrm{c}}_t, \mathbf{a}^{\mathrm{c}}_t),\,
           (\mathbf{h}^{\mathrm{f}}_t, \mathbf{a}^{\mathrm{f}}_t)
           \gets \mathcal{G}(\mathcal{C}_t)$ 
    \textit{\# Bi-level generation}
    \STATE $\mathbf{a}^{\star}_t \gets [\,\mathbf{a}^{\mathrm{c}}_t;\, \mathbf{a}^{\mathrm{f}}_t\,]$
    \STATE $\hat{\mathbf{A}}_{t:t+H} \gets \mathcal{D} (\mathbf{a}^{\star}_t)$
    \textit{\# Chunked action decoding}
\ENDFOR
\end{algorithmic}
\end{algorithm}

\subsection{The Base Vision-Language Model}
\label{subsec:basevla}
We build KineVLA on top of OpenVLA~\cite{kim2025openvla} as the base vision–language. Whereas OpenVLA originally adopts a discrete action tokenization scheme, we replace it with our proposed bi-level action representation. The action representation are learned via a residual VQ-VAE~\cite{zeghidour2021soundstream} trained for action reconstruction, enabling compact yet expressive control sequences. We retain OpenVLA’s Prismatic-7B~\cite{karamcheti2024prismatic} vision–language backbone, pre-trained on large-scale image–text data, and perform a second-stage adaptation using LoRA (rank = 32) applied to all linear layers, with Gaussian initialization of the LoRA weights. The input pipeline mirrors OpenVLA except that the natural-language instruction channel is replaced by fine-grained kinematics; other inputs (e.g., visual observations and task context) remain unchanged. The decoder jointly emits reason text and action tokens, both mapped into the same vocabulary index space as text, facilitating the use of a unified entropy loss. Unless otherwise noted, we adhere to OpenVLA’s standard data preprocessing, and optimization hyperparameters.

\begin{figure*}
    \centering
    \includegraphics[width=0.92\linewidth]{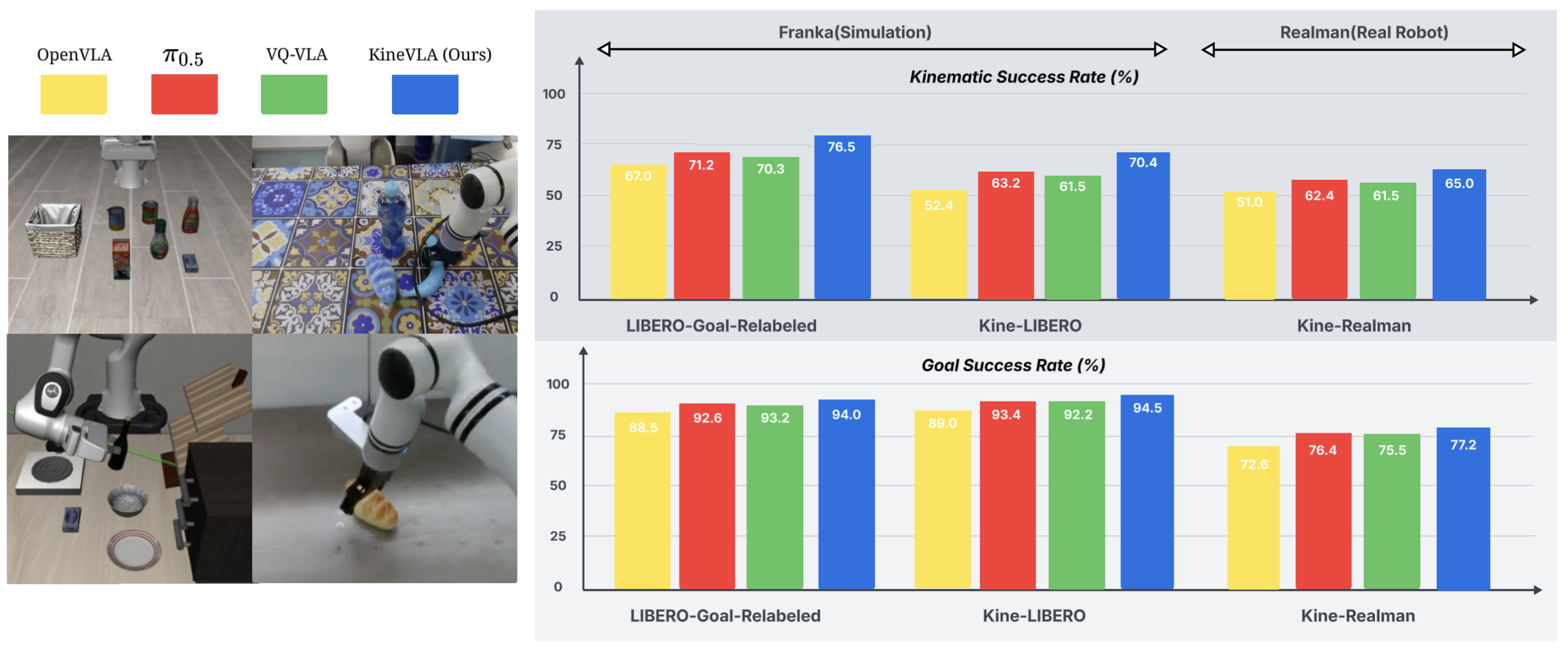}
    \caption{Experimental Results and Comparisons. We benchmark our method across the three proposed kinematics-aware datasets, encompassing both simulation and real-world robotic experiments. The left two figures illustrate example environments, while the bar chart on the right presents the goal and kinematic success rates.}
    \label{fig:res_3}
\end{figure*}

\section{Experiments}
\label{sec:experi}

We assess the effectiveness of our proposed approach through a series of experiments conducted on both simulation benchmarks and real-world robotic manipulation tasks. These experiments are designed to answer the following research questions:
\begin{itemize}
    \item Whether our method provides clear advantages over diffusion-based models and single-level VQ-VAE paradigms in kinematics-rich manipulation tasks.
    \item Does it achieve strong performance in both simulation and real-world robot settings?
    \item How much does the inclusion of chain-of-thought reasoning contribute to overall model performance?
    \item How effective is mutual information optimization in improving the prediction of reason text and action outputs?
\end{itemize}

\subsection{Experimental Setup}
\label{sec:experimental-setup}
We leverage three datasets (LIBERO-Goal-Relabeled, LIBERO-GOAL, Kine-Realman) to comprehensively cover a wide variety of goals and to annotate them with diverse, fine-grained kinematics-rich instructions. For details, please refer to Table~\ref{tab:datasets} in Appendix for a summary of our proposed dataset. Additional details are provided in the Appendix. All kinematics-related language descriptions in our datasets are defined in a consistent third-person camera coordinate frame aligned with human observation. Directional and positional terms such as left, right, near, and far are specified relative to the external camera view, rather than the robot base or end-effector frame. For each task, we curate a total of 50 demonstrations and assess performance across 20 evaluation trials. RVQ-VAE is trained on a single
A100 GPU with a batch size of 1024. For a fair comparison, all fine-tuning is conducted for 50K gradient steps with a total batch size of 64, using 4 A100 GPUs and an action chunk length of K=5.

\noindent \textbf{Evaluation Metrics.} For reasoning evaluation, we assess quality using three complementary metrics. BERTScore measures contextual semantic similarity and is robust to paraphrasing, while BLEU reports n-gram overlap as a surface-level metric. In addition, we include an Intervention metric, which replaces the predicted reasoning tokens with either randomly sampled reasoning tokens or mismatched reasoning tokens drawn from other instructions.

\noindent \textbf{Baselines.} We compare with OpenVLA~\cite{kim2025openvla}, $\pi_{0.5}$~\cite{intelligence2025pi_}, and VQ-VLA~\cite{wang2025vq}. OpenVLA is a vision–language–action transformer that conditions on image features and a text goal to regress continuous low-level actions. $\pi_{0.5}$ is a diffusion-based model, which uses co-training on heterogeneous tasks to enable broad generalization. VQ-VLA discretizes the action space using a single-level vector-quantized action tokenizer. For fairness, these methods consume the same vision observations and the same kinematics-rich language instructions.

\subsection{Evaluations Results}
\label{sec:eva-res}

\noindent \textbf{Quantivative Performance Analysis}. As shown in Figure ~\ref{fig:res_3}, all methods achieve comparable performance in terms of goal success rate across both simulation and real-world benchmarks, with only marginal differences between models. This indicates that existing VLA approaches are generally effective at capturing high-level task objectives. In contrast, when evaluated on kinematics success rate, the performance gap becomes substantially larger. While baseline methods exhibit a notable drop in kinematic accuracy, KineVLA consistently achieves significantly higher success rates across all datasets.

This discrepancy highlights a key limitation of prior approaches, which tend to collapse fine-grained kinematic variations once the task goal is reached. By contrast, KineVLA explicitly decouples goal-level intent from kinematics-level execution through its bi-level action representation and structured reasoning alignment, enabling precise adherence to instruction-level kinematic constraints. These results demonstrate that modeling kinematics as a first-class component is essential for kinematics-rich tasks, and that improvements in such settings cannot be achieved by goal-oriented modeling alone.

\begin{figure*}
    \centering
    \includegraphics[width=0.97\linewidth]{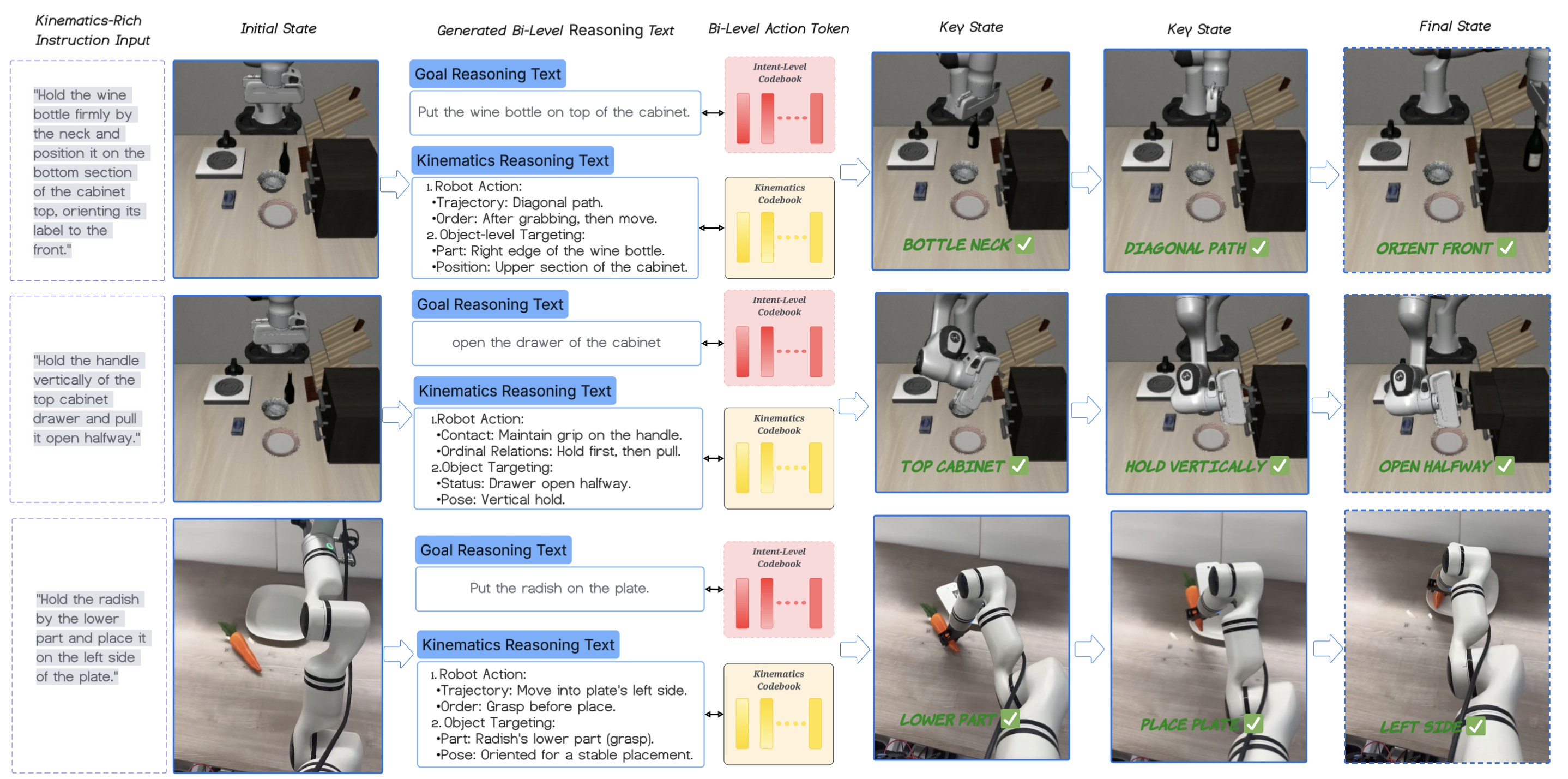}
    \caption{Task execution examples using KineVLA. From left to right, the figure shows the kinematics-rich instruction input, the initial state of the environment, followed by the bi-level reasoning text and action tokens generated by the KineVLA model. Next are two columns illustrating the key robot action states, and finally, the resulting final state after the robot executes the task.}
    \label{fig:examples3}
\end{figure*}

\noindent \textbf{Visualization Examples.} Figure~\ref{fig:examples3} illustrates qualitative examples of task execution using KineVLA, highlighting its ability to reason and act effectively in kinematics-aware manipulation tasks. From left to right, the figure demonstrates how KineVLA processes a kinematics-rich instruction, interprets the initial environmental state, and generates bi-level reasoning text, comprising goal-level and kinematics-level reasoning. The corresponding bi-level action tokens are then decoded into concrete robot motions, shown through key intermediate states and the final successful execution. This hierarchical reasoning framework enables KineVLA to bridge high-level semantic understanding with precise low-level motion control. For instance, it explicitly plans object contact, trajectory, and pose orientation, ensuring stability and accuracy during manipulation.

\noindent \textbf{Bi-level Reasoning Analysis.}
Table~\ref{tab:reasoning_bleu} shows that KineVLA achieves consistently high BLEU scores across all datasets, indicating strong alignment between the generated bi-level reasoning texts and the ground-truth annotations. To assess the necessity of bi-level reasoning, we conduct intervention-based analysis by replacing the predicted reasoning tokens with randomly sampled or mismatched ones while keeping visual inputs and task goals unchanged. This leads to a substantial drop in kinematics-following success rates from 76.5\% to 52.4\%, from 70.4\% to 48.6\%, and from 65.0\% to 42.4\% across the three datasets, respectively, while goal completion remains largely unaffected. These results demonstrate that bi-level reasoning tokens play a causal role in grounding kinematic constraints into action generation.

\noindent \textbf{Inference Speed Analysis.} As shown in Table \ref{tab:inference_time}, KineVLA does not introduce a significant inference overhead compared to single-level VQ-VAE models and diffusion-based methods. While slightly slower than VQ-VLA and $\pi_{0.5}$, the inference time of KineVLA remains in a similar range. This efficiency is achieved because the bi-level design operates at the feature and token level, requiring only a small increase in output tokens rather than additional decoding stages, resulting in minimal impact on overall inference speed.

\begin{table}[t]
\centering
\caption{Bi-level reasoning evaluation using the BLEU, BERTScore and Intervention metrics.}
\label{tab:reasoning_bleu}
\scriptsize
\begin{tabular}{lccc}
\toprule[1pt]
\textbf{Dataset} & \textbf{BLEU} & \textbf{BERTScore} & \textbf{Intervention (\%)}  \\
\midrule
LIBERO-Goal-Relabled  & 0.93 & 0.94 & 76.5$\rightarrow$52.4 \\
Kine-LIBERO   & 0.95 & 0.96 &70.4$\rightarrow$48.6 \\
Kine-Realman   & 0.92 & 0.94 &65.0$\rightarrow$42.4 \\
\bottomrule[1pt]
\end{tabular}

\end{table}

\begin{table}[t]
\centering
\caption{Inference Speed Comparison.}
\label{tab:inference_time}
\scriptsize
\begin{tabular}{cc}
\toprule[1pt]
\textbf{Methods}  & \textbf{ Inference Time }  \\
\midrule
OpenVLA   & 237ms  \\
$\pi_{0.5}$   & 86ms\\
VQ-VLA    & 84ms\\
KineVLA   & 96ms\\
\bottomrule[1pt]
\end{tabular}
\end{table}

\begin{table}[t]
\centering
\scriptsize
\caption{Ablation studies of KineVLA components. Success rates (\%) across two kinematics-rich datasets on LIBERO platform. “Baseline” refers to the standard vector-quantized action tokenization using OpenVLA.
“+ Bi-Rep” denotes the use of our Bi-Level action representation.
“+ Bi-Rea” indicates the incorporation of bi-level reasoning token.
“+ MI” refers to the addition of mutual information optimization between the reasoning text and the action.
 }
\label{tab:vla_bilevel}
\begin{tabular}{lcccc}
\toprule[1pt]
\textbf{Dataset} & \textbf{Baseline} & \textbf{+ Bi-Rep} & \textbf{+ Bi-Rea} & \textbf{+ MI}  \\
\midrule
LIBERO-Goal-Relabled  & 70.0 & 73.4 & 75.5 & 76.5 \\
Kine-LIBERO   & 61.5 & 66.8 & 68.6 & 70.4\\
\bottomrule[1pt]
\end{tabular}

\end{table}

\subsection{Ablation Study}
\label{sec:ablation}

The ablation results in Table 2 show the contribution of each proposed component in KineVLA. Starting from the \emph{Baseline}, which uses standard vector-quantized action tokenization with OpenVLA, introducing Bi-Rep leads to a clear improvement in performance, increasing success rates from 70.0\% to 73.4\% on LIBERO-Goal-Relabeled and from 61.5\% to 66.8\% on Kine-LIBERO. This demonstrates that the bi-level action representation effectively captures finer-grained motion semantics.

Adding \emph{Bi-Rea}, which introduces bi-level reasoning tokens, further improves performance to 75.5\% and 68.6\% on the two datasets. This demonstrates the importance of explicitly aligning language instructions with bi-level action representations, rather than relying on action modeling alone.

Finally, integrating \emph{MI}, the mutual information optimization between the reasoning text and the action, achieves the best results of 76.5\% on LIBERO-Goal-Relabeled and 70.4\% on Kine-LIBERO. This shows that aligning textual reasoning with action execution leads to more consistent and robust behavior.

Overall, these results confirm the effectiveness and complementarity of the proposed bi-level representation, reasoning, and optimization modules.

\section{Conclusion and Limitation}
\label{sec:conclusion}

In this work, we introduce KineVLA, a kinematics-rich vision–language–action framework that enables robots to understand and execute fine-grained motion instructions by explicitly disentangling kinematic sensitivity from goal invariance. This design allows a single goal to be realized through actions with varying kinematic granularity, improving both flexibility and interpretability. To support this task, we construct comprehensive simulated and real-world datasets with detailed kinematic annotations. Extensive experiments on tabletop manipulation demonstrate that KineVLA achieves superior precision and generalization compared to existing VLA models. While our current evaluation is limited to tabletop settings, the proposed bi-level formulation is naturally extensible to whole-body manipulation with more complex kinematic dependencies, which we leave for future work.

\bibliography{example_paper}
\bibliographystyle{icml2026}

\newpage
\appendix
\onecolumn

\section{Datasets and Visualization}

\noindent \textbf{LIBERO (Simulation).}
\label{subsec:libero}
Build on LIBERO platform~\cite{liu2023libero}, operators teleoperate the end-effector with a SpaceMouse. For each rollout, we record time-stamped RGB and wrist frames, joint states, forward-kinematics end-effector pose, and the action stream. We adopt the same preprocessing pipeline as in ~\cite{liu2023libero}, which includes: (1) removing idle segments from the trajectories, (2) resizing all images to a resolution of 256×256 pixels, and (3) rotating each image by 180 degrees. We build two splits:

\begin{itemize}
    \item \textbf{LIBERO-Goal-Relabeled:}
    LIBERO-GOAL covers 10 tasks and focuses on scenarios where the robot needs to continually acquire new knowledge about motions and behaviors.
    We retain the visual observations and action trajectories from LIBERO-GOAL and relabel the language instructions into their corresponding kinematics-rich versions based on human observation, making each action more specific.  
    For example, “put the wine bottle on the rack” is relabeled as “grab the wine bottle by the neck and place it at the front of the rack.”  
    More details can be found in the supplementary material.
    \item \textbf{Kine-LIBERO (ours):}
    We build upon the LIBERO platform and collect a larger set of kinematics-rich tasks in table manipulation scene, consisting 266K frames in total. Each episode is annotated with kinematics-rich language instructions and bi-level reasoning texts (goal and kinematics). As illustrated in Figure~\ref{fig:teaser_data}, our dataset features variations across the entire timeline, with diverse object parts, action constraints, and target relations.  
    Further details are provided in the supplementary material.
\end{itemize}

\noindent \textbf{Realman-75 (Real Robot).}
\label{subsec:realman}
We construct the \emph{Kine-Realman} dataset using a Realman-75 robotic arm equipped with a parallel gripper, a calibrated RGB camera observing a fixed tabletop workspace, and an additional wrist-mounted camera. The robot is teleoperated via puppet manipulation, and we collect data for seven tasks featuring rich, fine-grained kinematic descriptions. For example, one instruction specifies “grasp the lower half of the carrot and place it on the left side of the plate.” Further details about the tasks are provided in the supplementary material.

\begin{figure*}
    \centering
    \includegraphics[width=\linewidth]{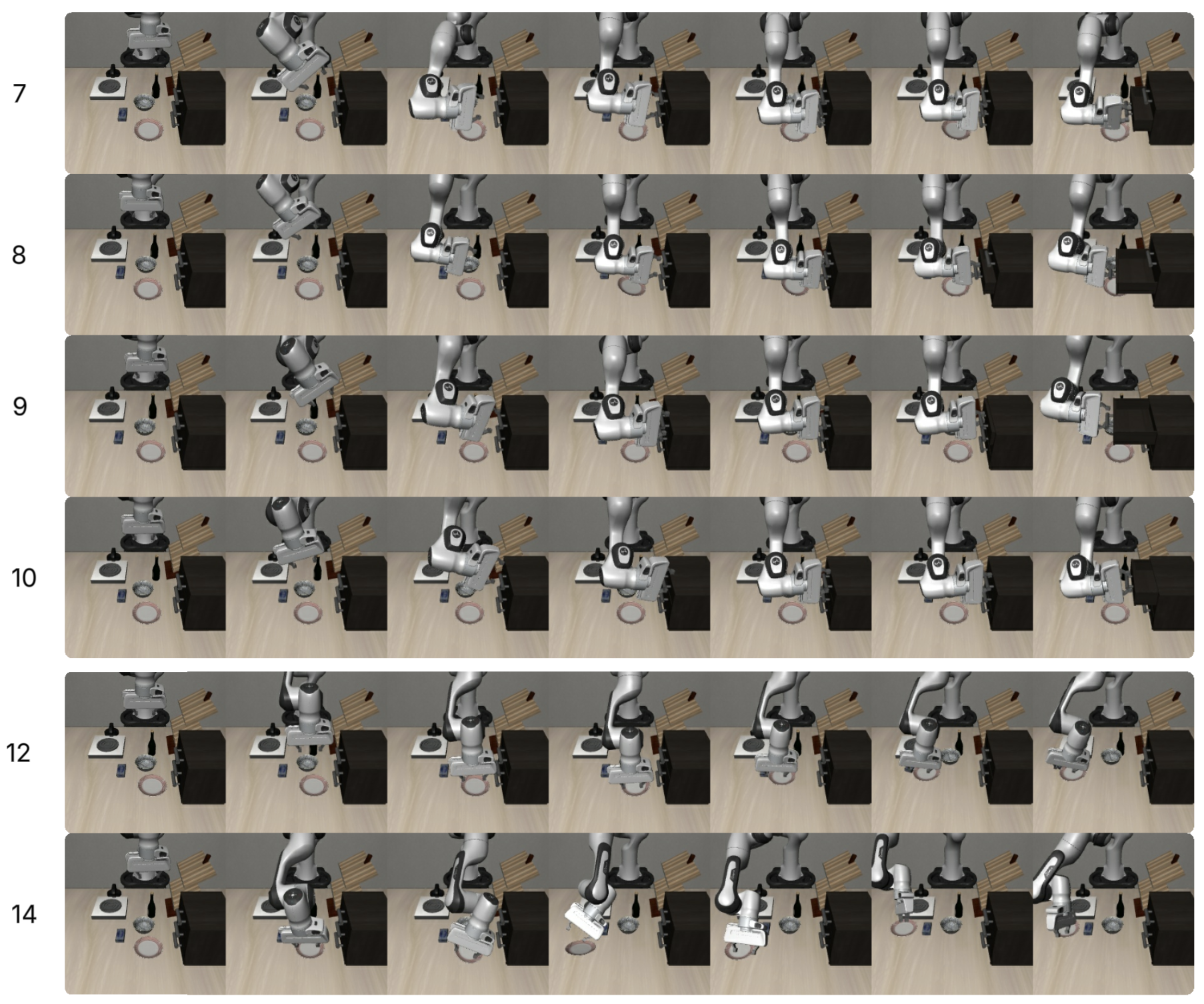}
    \caption{Execution results on the Kine-LIBERO dataset. The serial number corresponds to the entry in Table \ref{tab:kines}.}
    \label{fig:supp_kine_libero}
\end{figure*}

\begin{figure*}
    \centering
    \includegraphics[width=\linewidth]{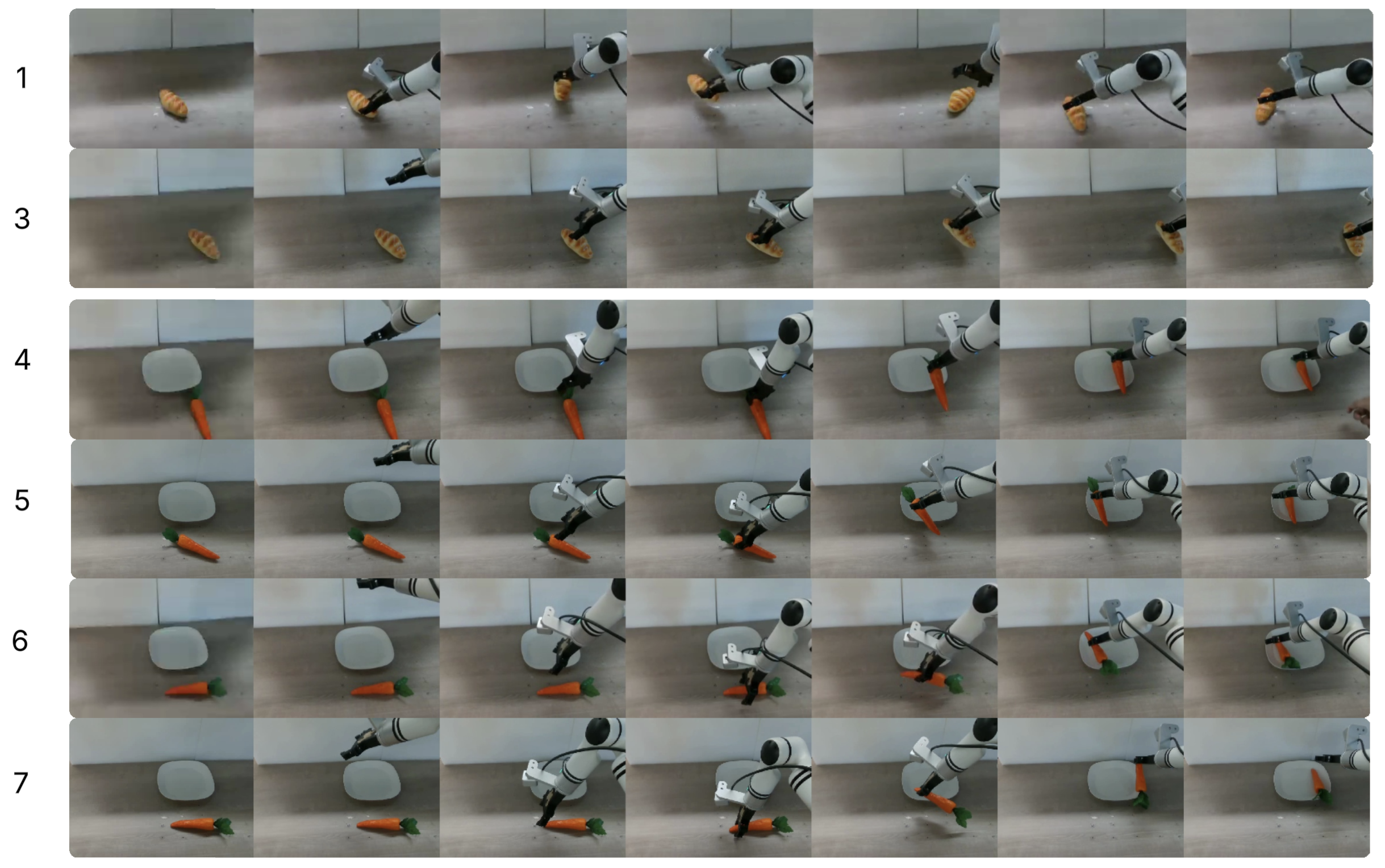}
    \caption{Execution results on the Kine-Realman-75 dataset. The serial number corresponds to the entry in Table \ref{tab:kine_realman}.}
    \label{fig:supp_kine_realman}
\end{figure*}

\noindent \textbf{Visualization.} As detailed in Tables \ref{tab:kine_realman},\ref{tab:kines}, and \ref{tab:libero-goal-labeled}, we have enumerated text examples from our annotated dataset alongside their corresponding bi-level Chain-of-Thought (CoT) annotations. Figure \ref{fig:supp_kine_libero} and \ref{fig:supp_kine_realman} illustrates the execution outcomes of our method on the Kine-LIBERO and Kine-Realman-75 benchmarks. The results from both benchmarks confirm that our method achieves precise execution of fine-grained kinematics operations in a consistent environment. {\bf Figure \ref{fig:supp_kine_libero}} demonstrates the executed tasks, categorized into two groups. The first group (Tasks 7-10) is concerned with opening drawers of different layers to various degrees (completely, halfway, slightly). The second group (Tasks 12,14) focuses on moving a plate following distinct trajectories (leftward then backward, diagonally toward the back-right). {\bf Figure \ref{fig:supp_kine_realman}} demonstrates the successful execution of two distinct task sets. The first set (Tasks 1, 3) demonstrates relocation of a piece of bread to different specified locations. The second set (Tasks 4-7) shows the grasping of a carrot at different parts (top/bottom) and its placement into different areas of a plate.

\begin{table*}[h]
\centering
\small
\caption{Summary of our proposed datasets collected via SpaceMouse and puppet manipulation. In each task, we gather 50 demonstrations.}
\label{tab:datasets}
\begin{tabular}{lcccccc}
\toprule[1pt]
\textbf{Split} & \textbf{Setting}  & \textbf{Frames} & \textbf{Language} & \textbf{Reasoning Text} & \textbf{Vision} & \textbf{Action Space} \\
\hline
LIBERO-Goal-Relabeled & Simulation & 63K &  Kinematics-rich & Bi-Level & Front, Wrist  & EEF Pose \\
Kine-LIBERO  & Simulation & 266K & Kinematics-rich & Bi-Level & Front, Wrist & EEF Pose \\
Kine-Realman-75   & Real robot & 49K & Kinematics-rich & Bi-Level & Front, Wrist & Joint Positions \\
\bottomrule[1pt]
\end{tabular}

\end{table*}

\onecolumn

\begingroup
\small
\setlength{\tabcolsep}{6pt}
\renewcommand{\arraystretch}{1.1}

\begin{longtable}{p{0.05\textwidth} p{0.28\textwidth} p{0.32\textwidth} p{0.32\textwidth}}
\caption{\bf LIBERO-Goal-Relabeled Dataset.} \label{tab:libero-goal-labeled} \\
\toprule
\textbf{No.} & \textbf{Kinematic language instruction} & \textbf{Coarse-grained CoT} & \textbf{Fine-grained CoT} \\
\midrule
\endfirsthead

\toprule
\textbf{No.} & \textbf{Kinematic language instruction} & \textbf{Coarse-grained CoT} & \textbf{Fine-grained CoT} \\
\midrule
\endhead

\midrule
\multicolumn{4}{r}{\textit{Continued on next page}} \\
\bottomrule
\endfoot

\bottomrule
\endlastfoot

1 & grasp the left edge of the bowl, move it forward, and place it precisely at the center of the plate & put the bowl on the plate & 1. Robot Action-level Constraint: Trajectory shape - forward linear movement; Contact timing - grasp then place; Ordinal relations - sequential grasp, move, place. 2. Object-level Targeting: Target - bowl's left edge for grasp, plate's center for placement; Pose/Facing/Angle - precise placement with implied alignment. \\
2 & grab the wine bottle by the neck and place it on the rack & put the wine bottle on the rack & 1. Robot Action-level Constraint: Contact Timing - Must initiate/break contact (grab/release). Trajectory Shape - Requires transfer motion from source (neck) to target (rack). Ordinal Relations - Strict sequence: grab before place. 2. Object-level Targeting: Grasp Target - The neck (specific part) of wine bottle. Place Target - The rack (specific object); final pose constrained by rack structure (must be placed on it). \\
3 & press the handle to open the top drawer to its maximum extension, then grasp the left edge of the bowl and place it in the center of the drawer & open the top drawer and put the bowl inside & 1. Robot Action-level Constraint: State/Trajectory: Press, then grasp and place.  Contact Timing: Press handle first, then grasp bowl after drawer is open.  Ordinal Relations: Sequential: open drawer, then grasp bowl, then place bowl. 2. Object-level Targeting: Handle/Surface/Edge: Handle (of drawer), left edge (of bowl).  Pose/Facing/Angle: Maximum extension (drawer state), center (target placement pose). \\
4 & grab the cream cheese by the middle and place it in the center of the bowl & put the cream cheese in the bowl & 1. Robot Action-level Constraint: Trajectory shape - Straight-line placement motion; Contact timing - Grasp first, then place; Ordinal relations - Sequential (grab, then place). 2.Object-level Targeting: Handle/Surface - Target middle of object (cream cheese); Pose/Facing - Orientation-agnostic grasp; Placement - Center of bowl (specific positional target). \\
5 & hold the wine bottle firmly by the neck and position it on the upper section of the cabinet top, orienting the bottle so that its label is facing to the left & put the wine bottle on top of the cabinet & 1. Robot Action-level Constraint: State/Trajectory: Hold firmly, position on upper section. Contact Timing: Grip neck before positioning. Ordinal Relations: Orient after positioning. 2. Object-level Targeting: Handle/Edge/Surface: Neck (grip), cabinet top (placement). Pose/Facing/Angle: Label facing left. \\
6 & place the end-effector at the center of the plate and move it leftward then backward along the table plane to reposition the plate about close in front of the stove & push the plate to the front of the stove & 1. Robot Action-level Constraint: State/Trajectory: Move end-effector leftward then backward along the table plane. Contact Timing: Maintain contact with the plate. Ordinal Relations: Sequential motion: first leftward, then backward. 2. Object-level Targeting: Target Part: Center of the plate (initial contact). Pose/Facing/Angle: Align end-effector to plate center; movement constrained to table plane. Final Placement: close in front of the stove. \\
7 & move left to the stove knob, then turn it clockwise to ignite & turn on the stove & Robot Action-level Constraint: Trajectory shape - Linear translation (move left) followed by rotational (turn clockwise). Contact timing - Sequential (move then turn). Ordinal relations - Strict order: translation before rotation. Object-level Targeting: Target object part - Stove knob (specific component). Pose/facing/angle - Clockwise rotation (specific angular direction). \\
8 & grab the left edge of this bowl and move it along a diagonal path to the center of the stove & put the bowl on the stove & Robot Action-level Constraint: Trajectory shape - Diagonal path; Contact timing - Grab (initial contact) Object-level Targeting: Target part - Left edge; Object - Bowl; Final pose - Center of the stove. \\
9 & grab the right edge of this bowl and move it along a diagonal path to the upper section of the cabinet & put the bowl on top of the cabinet & 1. Robot Action-level Constraint: Trajectory shape - Diagonal path; Contact timing - After grabbing, then move 2. Object-level Targeting: Target component - Right edge of the bowl; Spatial relation - Upper section of the cabinet. \\
10 & hold the handle of the middle cabinet drawer and pull it open completely & open the middle drawer of the cabinet & 1. Robot Action-level Constraint: Contact Timing: Initiate contact Trajectory Shape: Linear pulling Ordinal Relations: Grasp then pull 2. Object-level Targeting: Target Part: Middle cabinet drawer handle Action on Target: Pull open completely. \\
\end{longtable}
\endgroup

\begin{longtable}{p{0.8cm} p{5.2cm} p{5.2cm} p{5.2cm}}
\caption{\bf Kine-LIBERO Dataset}\label{tab:kines}\\
\toprule
\textbf{No.} & \textbf{Kinematic language instruction} & \textbf{Coarse-grained CoT} & \textbf{Fine-grained CoT} \\
\midrule
\endfirsthead

\toprule
\textbf{No.} & \textbf{Kinematic language instruction} & \textbf{Coarse-grained CoT} & \textbf{Fine-grained CoT} \\
\midrule
\endhead

\midrule
\multicolumn{4}{r}{\small\itshape Continued on next page}\\
\bottomrule
\endfoot

\bottomrule
\endlastfoot
1 & grab the right edge of this bowl and move it along a diagonal path to the upper section of the cabinet & put the bowl on top of the cabinet & 1. Robot Action-level Constraint: Trajectory shape - Diagonal path; Contact timing - After grabbing, then move 2. Object-level Targeting: Target component - Right edge of the bowl; Spatial relation - Upper section of the cabinet \\
2 & grab the left edge of this bowl and move it along a diagonal path to the upper section of the cabinet & put the bowl on top of the cabinet & 1. Robot Action-level Constraint: Trajectory shape - Diagonal path; Contact timing - After grabbing, then move 2. Object-level Targeting: Target component - Left edge of the bowl; Spatial relation - Upper section of the cabinet \\
3 & grab the right edge of this bowl and move it along a diagonal path to the bottom section of the cabinet & put the bowl on top of the cabinet & 1. Robot Action-level Constraint: Trajectory shape - Diagonal path; Contact timing - After grabbing, then move 2. Object-level Targeting: Target component - Right edge of the bowl; Spatial relation - Bottom section of the cabinet \\
4 & grab the left edge of this bowl and move it along a diagonal path to the bottom section of the cabinet & put the bowl on top of the cabinet & 1. Robot Action-level Constraint: Trajectory shape - Diagonal path; Contact timing - After grabbing, then move 2. Object-level Targeting: Target component - Left edge of the bowl; Spatial relation - Bottom section of the cabinet \\
5 & hold the handle of the middle cabinet drawer and pull it open completely & open the middle drawer of the cabinet & 1. Robot Action-level Constraint: Contact Timing: Initiate contact Trajectory Shape: Linear pulling. Ordinal Relations: Grasp then pull 2. Object-level Targeting: Target Part: Middle cabinet drawer handle. Action on Target: Pull open completely \\
6 & vertically hold the handle of the middle cabinet drawer and pull it open completely & open the middle drawer of the cabinet & 1. Robot Action-level Constraint: Contact Timing: Initiate contact. Trajectory Shape: Linear pulling. Ordinal Relations: Grasp then pull. 2. Object-level Targeting: Target Part: Middle cabinet drawer handle. Action on Target: Pull open completely \\
7 & vertically hold the handle of the middle cabinet drawer and pull it open halfway & open the middle drawer of the cabinet & 1. Robot Action-level Constraint: Contact Timing: Initiate contact. Trajectory Shape: Linear pulling. Ordinal Relations: Grasp then pull. 2. Object-level Targeting: Target Part: Middle cabinet drawer handle. Action on Target: Pull open halfway \\
8 & vertically hold the handle of the middle cabinet drawer and pull it slightly open & open the middle drawer of the cabinet & 1. Robot Action-level Constraint: Contact Timing: Initiate contact. Trajectory Shape: Linear pulling. Ordinal Relations: Grasp then pull. 2. Object-level Targeting: Target Part: Middle cabinet drawer handle. Action on Target: Pull open slightly \\
9 & vertically hold the handle of the top cabinet drawer and pull it open completely & open the top drawer of the cabinet & 1. Robot Action-level Constraint: Contact Timing: Initiate contact. Trajectory Shape: Linear pulling. Ordinal Relations: Grasp then pull. 2. Object-level Targeting: Target Part: Top cabinet drawer handle. Action on Target: Pull open completely \\
10 & vertically hold the handle of the top cabinet drawer and pull it open halfway & open the top drawer of the cabinet & 1. Robot Action-level Constraint: Contact Timing: Initiate contact. Trajectory Shape: Linear pulling. Ordinal Relations: Grasp then pull. 2. Object-level Targeting: Target Part: Top cabinet drawer handle. Action on Target: Pull open halfway \\
11 & vertically hold the handle of the top cabinet drawer and pull it slightly open & open the top drawer of the cabinet & 1. Robot Action-level Constraint: Contact Timing: Initiate contact. Trajectory Shape: Linear pulling. Ordinal Relations: Grasp then pull. 2. Object-level Targeting: Target Part: Top cabinet drawer handle. Action on Target: Pull open slightly \\
12 & push the plate and move it leftward then backward along the table plane to reposition the plate about close in front of the stove & push the plate to the front of the stove & 1. Robot Action-level Constraint: State/Trajectory: Move end-effector leftward then backward along the table plane. Contact Timing: Maintain contact with the plate. Ordinal Relations: Sequential motion: first leftward, then backward. 2. Object-level Targeting: Target Part: Center of the plate (initial contact). Pose/Facing/Angle: Align end-effector to plate center; movement constrained to table plane. Final Placement: \textasciitilde{}close in front of the stove. \\
13 & push the plate and move it rightward then backward along the table plane to reposition the plate about far in front of the stove & push the plate to the front of the stove & 1. Robot Action-level Constraint: State/Trajectory: Move end-effector rightward then backward along the table plane. Contact Timing: Maintain contact with the plate. Ordinal Relations: Sequential motion: first rightward, then backward. 2. Object-level Targeting: Target Part: Center of the plate (initial contact). Pose/Facing/Angle: Align end-effector to plate center; movement constrained to table plane. Final Placement: \textasciitilde{}far in front of the stove. \\
14 & push the plate and move it diagonally toward the back-right along the table plane to reposition the plate about close in front of the stove & push the plate to the front of the stove & 1. Robot Action-level Constraint: State/Trajectory: Move end-effector diagonally toward back-right along the table plane. Contact Timing: Maintain contact with the plate. Ordinal Relations: Sequential motion: first diagonally back-right. 2. Object-level Targeting: Target Part: Center of the plate (initial contact). Pose/Facing/Angle: Align end-effector to plate center; movement constrained to table plane. Final Placement: \textasciitilde{}close in front of the stove. \\
15 & push the plate and move it diagonally toward the back-right along the table plane to reposition the plate about far in front of the stove & push the plate to the front of the stove & 1. Robot Action-level Constraint: State/Trajectory: Move end-effector diagonally toward back-right along the table plane. Contact Timing: Maintain contact with the plate. Ordinal Relations: Sequential motion: first diagonally back-right. 2. Object-level Targeting: Target Part: Center of the plate (initial contact). Pose/Facing/Angle: Align end-effector to plate center; movement constrained to table plane. Final Placement: \textasciitilde{}far in front of the stove. \\
16 & push the plate and move it rightward then backward along the table plane to reposition the plate about close in right of the stove & push the plate to the right the stove & 1. Robot Action-level Constraint: State/Trajectory: Move end-effector rightward then backward along the table plane. Contact Timing: Maintain contact with the plate. Ordinal Relations: Sequential motion: first rightward, then backward. 2. Object-level Targeting: Target Part: Center of the plate (initial contact). Pose/Facing/Angle: Align end-effector to plate center; movement constrained to table plane. Final Placement: \textasciitilde{}close in right of the stove. \\
17 & push the plate and move it diagonally toward the back-right along the table plane to reposition the plate about close in right of the stove & push the plate to the right the stove & 1. Robot Action-level Constraint: State/Trajectory: Move end-effector diagonally toward back-right along the table plane. Contact Timing: Maintain contact with the plate. Ordinal Relations: Sequential motion: first diagonally back-right.2. Object-level Targeting: Target Part: Center of the plate (initial contact). Pose/Facing/Angle: Align end-effector to plate center; movement constrained to table plane. Final Placement: \textasciitilde{}close in right of the stove. \\
18 & grab the wine bottle by the neck and place it on the rack & put the wine bottle on the rack & 1. Robot Action-level Constraint: Contact Timing - Must initiate/break contact (grab/release). Trajectory Shape - Requires transfer motion from source (neck) to target (rack). Ordinal Relations - Strict sequence: grab before place. 2. Object-level Targeting: Grasp Target - The neck (specific part) of wine bottle. Place Target - The rack (specific object); final pose constrained by rack structure (must be placed on it). \\
19 & grab the wine bottle by the body and place it at the front of the rack & put the wine bottle on the rack & 1. Robot Action-level Constraint: Contact Timing - Must initiate/break contact (grab/release). Trajectory Shape - Requires transfer motion from source (body) to target (rack front). Ordinal Relations - Strict sequence: grab before place. 2. Object-level Targeting: Grasp Target - The body (specific part) of wine bottle. Place Target - The front of the rack (specific location); final pose constrained by rack structure (must be placed on it). \\
20 & grab the wine bottle by the neck and place it at the back of the rack & put the wine bottle on the rack & 1. Robot Action-level Constraint: Contact Timing - Must initiate/break contact (grab/release). Trajectory Shape - Requires transfer motion from source (neck) to target (rack back). Ordinal Relations - Strict sequence: grab before place. 2. Object-level Targeting: Grasp Target - The neck (specific part) of wine bottle. Place Target - The back of the rack (specific location); final pose constrained by rack structure (must be placed on it). \\
21 & grab the wine bottle by the body and place it at the back of the rack & put the wine bottle on the rack & 1. Robot Action-level Constraint: Contact Timing - Must initiate/break contact (grab/release). Trajectory Shape - Requires transfer motion from source (body) to target (rack back). Ordinal Relations - Strict sequence: grab before place. 2. Object-level Targeting: Grasp Target - The body (specific part) of wine bottle. Place Target - The back of the rack (specific location); final pose constrained by rack structure (must be placed on it). \\
22 & grab the right edge of this bowl and move it along a diagonal path to the center of the stove & put the bowl on the stove & Robot Action-level Constraint: Trajectory shape - Diagonal path; Contact timing - Grab (initial contact) Object-level Targeting: Target part - Right edge; Object - Bowl; Final pose - Center of the stove \\
23 & grab the left edge of this bowl and move it along a diagonal path to the center of the stove & put the bowl on the stove & Robot Action-level Constraint: Trajectory shape - Diagonal path; Contact timing - Grab (initial contact). Object-level Targeting: Target part - Left edge; Object - Bowl; Final pose - Center of the stove \\
24 & hold the wine bottle firmly by the neck and position it on the upper section of the cabinet top, orienting the bottle so that its label is facing to the left & put the wine bottle on top of the cabinet & 1. Robot Action-level Constraint: State/Trajectory: Hold firmly, position on upper section. Contact Timing: Grip neck before positioning. Ordinal Relations: Orient after positioning. 2. Object-level Targeting: Handle/Edge/Surface: Neck (grip), cabinet top (placement). Pose/Facing/Angle: Label facing left. \\
25 & hold the wine bottle firmly by the neck and position it on the bottom section of the cabinet top, orienting the bottle so that its label is facing to the right & put the wine bottle on top of the cabinet & 1. Robot Action-level Constraint: State/Trajectory: Hold firmly, position on lower section. Contact Timing: Grip neck before positioning. Ordinal Relations: Orient after positioning. 2. Object-level Targeting: Handle/Edge/Surface: Neck (grip), cabinet top (placement). Pose/Facing/Angle: Label facing right. \\
26 & hold the wine bottle firmly by the neck and position it on the upper section of the cabinet top, orienting the bottle so that its label is facing to the left & put the wine bottle on top of the cabinet & 1. Robot Action-level Constraint: State/Trajectory: Hold firmly, position on upper section. Contact Timing: Grip neck before positioning. Ordinal Relations: Orient after positioning. 2. Object-level Targeting: Handle/Edge/Surface: Neck (grip), cabinet top (placement). Pose/Facing/Angle: Label facing left. \\
27 & hold the wine bottle firmly by the neck and position it on the bottom section of the cabinet top, orienting the bottle so that its label is facing to the left & put the wine bottle on top of the cabinet & 1. Robot Action-level Constraint: State/Trajectory: Hold firmly, position on lower section. Contact Timing: Grip neck before positioning. Ordinal Relations: Orient after positioning. 2. Object-level Targeting: Handle/Edge/Surface: Neck (grip), cabinet top (placement). Pose/Facing/Angle: Label facing left. \\
28 & hold the wine bottle firmly by the neck and position it on the upper section of the cabinet top, orienting the bottle so that its label is facing to the front & put the wine bottle on top of the cabinet & 1. Robot Action-level Constraint: State/Trajectory: Hold firmly, position on upper section. Contact Timing: Grip neck before positioning. Ordinal Relations: Orient after positioning. 2. Object-level Targeting: Handle/Edge/Surface: Neck (grip), cabinet top (placement). Pose/Facing/Angle: Label facing front. \\
29 & hold the wine bottle firmly by the neck and position it on the bottom section of the cabinet top, orienting the bottle so that its label is facing to the front & put the wine bottle on top of the cabinet & 1. Robot Action-level Constraint: State/Trajectory: Hold firmly, position on lower section. Contact Timing: Grip neck before positioning. Ordinal Relations: Orient after positioning. 2. Object-level Targeting: Handle/Edge/Surface: Neck (grip), cabinet top (placement). Pose/Facing/Angle: Label facing front. \\
30 & hold the wine bottle firmly by the neck and position it on the upper section of the cabinet top, orienting the bottle so that its label is facing to the back & put the wine bottle on top of the cabinet & 1. Robot Action-level Constraint: State/Trajectory: Hold firmly, position on upper section. Contact Timing: Grip neck before positioning. Ordinal Relations: Orient after positioning. 2. Object-level Targeting: Handle/Edge/Surface: Neck (grip), cabinet top (placement). Pose/Facing/Angle: Label facing back. \\
31 & hold the wine bottle firmly by the neck and position it on the bottom section of the cabinet top, orienting the bottle so that its label is facing to the back & put the wine bottle on top of the cabinet & 1. Robot Action-level Constraint: State/Trajectory: Hold firmly, position on lower section. Contact Timing: Grip neck before positioning. Ordinal Relations: Orient after positioning. 2. Object-level Targeting: Handle/Edge/Surface: Neck (grip), cabinet top (placement). Pose/Facing/Angle: Label facing back. \\
\end{longtable}

\begin{table*}[htbp]
\centering
\setlength{\tabcolsep}{4pt}
\renewcommand{\arraystretch}{1.1}
\caption{\bf Kine-Realman Dataset}
\label{tab:kine_realman}
\begin{tabular}{p{0.04\linewidth} p{0.28\linewidth} p{0.32\linewidth} p{0.32\linewidth}}
\toprule
\textbf{No.} & \textbf{Kinematic language instruction} & \textbf{Coarse-grained CoT} & \textbf{Fine-grained CoT} \\
\midrule

1 & pick up the bread and move it to the left front & move the bread & Robot Action: Trajectory Shape - To the left front; Ordinal Relations - Pick first, then move. Object Targeting: Target Object - The bread; Spatial Relation - Move it to a location (left front). \\
2 & pick up the bread and move it to the right front & move the bread & Robot Action: Trajectory Shape - To the right front; Ordinal Relations - Pick first, then move. Object Targeting: Target Object - The bread; Spatial Relation - Move it to a location (right front). \\
3 & pick up the bread and move it to the front & move the bread & Robot Action: Trajectory Shape - To the front; Ordinal Relations - Pick first, then move. Object Targeting: Target Object - The bread; Spatial Relation - Move it to a location (front). \\
4 & hold the carrot by the upper part and place it on the right side of the plate & place the carrot in the plate & Robot Action-level Constraint: State/Trajectory - Move carrot from held to placement pose. Ordinal Relations - Grasp first, then place. Object-level Targeting: Carrot - Target handle (upper part). Plate - Target surface (top side). Spatial Relation - Final pose on the right side of the plate. \\
5 & hold the carrot by the upper part and place it on the left side of the plate & place the carrot in the plate & Robot Action-level Constraint: State/Trajectory - Move carrot from held to placement pose. Ordinal Relations - Grasp first, then place. Object-level Targeting: Carrot - Target handle (upper part). Plate - Target surface (top side). Spatial Relation - Final pose on the left side of the plate. \\
6 & hold the carrot by the lower part and place it on the left side of the plate & place the carrot in the plate & Robot Action-level Constraint: State/Trajectory - Move carrot from held to placement pose. Ordinal Relations - Grasp first, then place. Object-level Targeting: Carrot - Target handle (lower part). Plate - Target surface (top side). Spatial Relation - Final pose on the left side of the plate. \\
7 & hold the carrot by the lower part and place it on the right side of the plate & place the carrot in the plate & Robot Action-level Constraint: State/Trajectory - Move carrot from held to placement pose. Ordinal Relations - Grasp first, then place. Object-level Targeting: Carrot - Target handle (lower part). Plate - Target surface (top side). Spatial Relation - Final pose on the right side of the plate. \\
\bottomrule
\end{tabular}
\end{table*}

\end{document}